\documentclass[11pt]{article}
\usepackage{geometry}
\usepackage{coling2020}
\usepackage{times}
\usepackage{url}
\usepackage{hyperref}
\usepackage{latexsym}
\usepackage{microtype}
\usepackage{nert}

\usepackage{subcaption}

\colingfinalcopy 

\title{Team DoNotDistribute at SemEval-2020 Task 11: Features, Finetuning, and Data Augmentation in Neural Models for Propaganda Detection in News Articles
}

\author{Michael Kranzlein, Shabnam Behzad, Nazli Goharian\\
    Department of Computer Science\\
        Georgetown University\\
        \{\emldisplay{mmk119@georgetown.edu}{mmk119}, \emldisplay{sb1796@georgetown.edu}{sb1796}\}\texttt{@georgetown.edu}\\
        \emldisplay{nazli@ir.cs.georgetown.edu}{nazli}\texttt{@ir.cs.georgetown.edu}\\
}

\date{}

\begin{document}
\maketitle
\begin{abstract}
  This paper presents our systems for SemEval 2020 Shared Task 11: Detection of Propaganda Techniques in News Articles. We participate in both the span identification and technique classification subtasks and report on experiments using different BERT-based models along with handcrafted features. Our models perform well above the baselines for both tasks, and we contribute ablation studies and discussion of our results to dissect the effectiveness of different features and techniques with the goal of aiding future studies in propaganda detection.
\end{abstract}

\section{Introduction}

Propagandist news articles are misleading in nature and aim at biasing their audience towards a particular point of view by using psychological and rhetorical techniques, including loaded language, name calling, repetition, exaggeration, minimization, etc. With the rapid growth in the number of online sources of information and the speed with which information spreads online, manual flagging of propagandist news articles has become untenable, leading to an ongoing need for new research on methods for identifying these articles automatically to mitigate the negative influence they might have on users.

Until very recently, most of the work in this area has focused on article-level detection~\cite{rashkin-etal-2017-truth,barron2019proppy}. However, in 2019, Da San Martino et al.~\shortcite{da-san-martino-etal-2019-fine} published a corpus of English news articles with individual spans of propaganda annotated that addresses the problem at a more granular level. This corpus is used in shared tasks at NLP4IF-2019~\cite{da-san-martino-etal-2019-findings} and at SemEval-2020~\cite{DaSanMartinoSemeval20task11}. The 2020 shared task is a modified version of the prior year's task and includes two subtasks\footnote{\url{https://propaganda.qcri.org/semeval2020-task11/}}:
\begin{enumerate}
    \item \textbf{Span Identification (SI):} Given a plain-text document, identify those specific fragments which contain at least one propaganda technique. This is a binary sequence tagging task.
    
    \item \textbf{Technique Classification (TC):} Given a text fragment identified as propaganda and its document context, identify the applied propaganda technique in the fragment.
\end{enumerate}

We present our models\footnote{Model hyperparameters, code, and instructions for reproducing our results are available at \url{https://github.com/mkranzlein/propaganda}} for both tasks alongside discussions of our results and ablations.

\section{Background}

Automatic fact-checking and propaganda detection in news articles have attracted considerable attention in recent years~\cite{potthast-etal-2018-stylometric,helmstetter2018weakly,baly-etal-2018-predicting}. Many of these works focus on identifying propaganda at the article level.\ Rashkin et al.~\shortcite{rashkin-etal-2017-truth} created a corpus of news articles with 4 different news types: Trusted, Satire, Hoax and Propaganda. They compared the language of these different types and reported classification results using LSTM, Max-Ent, and Naive Bayes models. Barr{\'o}n-Cede\~no et al.~\shortcite{BARRONCEDENO20191849} experimented on this corpus with a maximum entropy classifier to discriminate propagandist from non-propagandist articles.

Da San Martino et al.~\shortcite{da-san-martino-etal-2019-fine} formulated the problem of detecting propaganda techniques in text and released a new corpus for the problem. They also presented a multi-granularity neural network for this task that outperformed several baselines. This dataset was used in the NLP4IF-2019 Shared Task~\cite{da-san-martino-etal-2019-findings} with 2 subtasks: Fragment-Level Classification (FLC) and Sentence-Level Classification (SLC). Multiple systems using novel approaches were submitted.

Gupta et al.~\shortcite{gupta-etal-2019-neural} designed multi-granularity and multi-tasking neural architectures which could jointly detect sentence- and fragment-level propaganda. They also report results on different ensemble schemes. Fadel et al.~\shortcite{fadel-etal-2019-pretrained} presented an ensemble of the transformer version of the Universal Sentence Encoder~\cite{cer2018universal} and BERT. Many other teams also incorporated BERT-base or BERT-large with a linear layer or LSTM~\cite{hou-chen-2019-caunlp,hua-2019-understanding,mapes-etal-2019-divisive,yoosuf-yang-2019-fine}. Li et al.~\shortcite{li-etal-2019-detection} tested a model that makes indirect use of BERT, but ultimately attained slightly higher performance using a logistic regression model incorporating TF-IDF vectors and emotion features, among others.

Several other teams also experimented with handcrafted features such as part-of-speech tags, character n-grams, punctuation frequency, sentence length variance, average word length and sentiment features~\cite{vlad-etal-2019-sentence,ferreira-cruz-etal-2019-sentence,al-omari-etal-2019-justdeep}.

The datasets for both 2019 subtasks are imbalanced. Almost all teams made some attempt to address this issue. Methods included undersampling, data augmentation techniques such as oversampling and synonym insertion, dropping words~\cite{yoosuf-yang-2019-fine,aggarwal-sadana-2019-nsit,tayyar-madabushi-etal-2019-cost,ek-ghanimifard-2019-synthetic} or adjusting the threshold for prediction probabilities~\cite{alhindi-etal-2019-fine}.

Here, we describe our efforts for both subtasks of SemEval 2020 task 11. For SI, we experimented with 3 different model architectures and combinations of features like BERT embeddings, part of speech (POS) tags, and named entities. For TC, our best model is an LSTM with BERT embeddings plus similar features. We were able to enhance the performance of this model via language model finetuning and data augmentation.

\section{Task 1: Span Identification}

Span identification shares some structural similarities with question answering (QA) and named entity recognition (NER), though these tasks all differ in purpose. In QA, many datasets contain question-context pairs with \textit{one} gold answer span highlighted within the context paragraph of each pair. In SI and NER, while we still need to highlight a portion of the text, there is no question/query, and there's no prespecified number of markables. In one sense, SI is more challenging than NER because propaganda spans show greater variance in length and content. Named entities are usually noun-like and fairly short, while the length of a propaganda span can range from one word—as is often the case for loaded language—to multiple sentences. On the other hand, SI is simpler in that the prediction is not multiclass. In this task's framing of the problem, predicting propaganda types is done separately in TC after the spans have already been marked.

\subsection{Preprocessing}
We built three types of models, which all rely on similar preprocessing steps. Each news article in the corpus is a sentence-segmented text file. Gold labels for propaganda spans are given as pairs of character start and stop indexes. This leads to an important design decision: do we look for propaganda at the token level or at the character level? While character-level predictions have been used to good effect in similar tasks like named entity recognition \cite{kuru-etal-2016-charner}, we choose to make predictions over tokens in order to make use of BERT's contextualized word embeddings. This does bound performance slightly, since some gold propaganda spans include partial tokens, and our models can only make decisions about entire tokens. Future models may benefit from combining token-level and character-level information.

Because gold labels are character-indexed but we want to predict token labels, our first preprocessing step is to tokenize and and generate a binary \textit{propaganda}~or~\textit{not propaganda} label for each token based on whether any of the token's characters fall inside of a character-indexed propaganda span. This requires knowing the character index of each token. We use spaCy, which provides this functionality in its tokenizer. However, this means we lose some performance in BERT since we're not using the wordpiece tokenizer, where converting character indexes to token indexes and vice-versa is much more challenging. 

For featurization, we use spaCy to automatically extract part-of-speech and named entity vectors. Then, in the training data, we count the number of propaganda spans consisting of a single token. For each token in our model input, we include the number of times it appeared in the training data as a single-token propaganda span as an additional feature. This is the propaganda keyword frequencies (KW) feature referenced in \cref{tab:si_dev_results}. We then get BERT embeddings (using the base version due to GPU memory constraints) for each token.

\subsection{Models}
\label{sec:si_models}
For each of our three SI models, a training instance is based on the tokens of one sentence. Each model makes a binary prediction about all the tokens in the sentence, and we convert these token predictions back to the original character-indexed format. For each article, we combine the predictions for each of the article's sentences and merge overlapping spans before evaluating.

\paragraph{BERT}
Our first model is the most straightforward. We simply train Huggingface's BertForTokenClassification off-the-shelf model on our preprocessed tokens and labels. This model is just the original transformer-based BERT architecture adapted for tasks like NER and SI by adding a linear layer that makes predictions over each token. 

\paragraph{LSTM} Our second model is LSTM-based and uses fixed BERT embeddings plus a feature vector of part-of-speech tags, named entities, and propaganda keyword frequencies.  

\paragraph{Bert Predictions \& Features}
Our third (and best) model integrates the benefits of the previous two models. This model does not use any embeddings directly. Instead, it uses the same feature vector as the LSTM model plus one additional feature for each token: the predicted probability of that token being propaganda according to the BERT model.

\subsection{Results}
Here, we report results and ablations on the dev set\footnote{Our dev set results and ablations reflect bug fixes and optimizations after and are slightly higher than our leaderboard results. On the test set, however, in order to remain consistent with the final rankings, we only report our leaderboard results.}. The LSTM model input consists of BERT embeddings and features, which include part-of-speech tags (POS), named entities (NER), and propaganda keyword frequencies (KW). Included ablations show how performance is affected by removing each feature individually (e.g. LSTM - POS) and by omitting all features (i.e. input of fixed BERT embeddings only). 

\begin{table}[h]
    \centering\small\setlength\tabcolsep{5pt}
    \begin{tabular}{l | c c c}
        &\textbf{P}&\textbf{R}&\textbf{F}\\
        \hline
        \textbf{\# Gold Spans} & \multicolumn{3}{c}{941}\\
        \hline

        LSTM (Embeddings only) & 25.3 & 46.5 & 32.8\\
        LSTM (POS + NER + KW) & 27.4 & 47.0 & 34.6\\
        LSTM - POS & 28.3 & 41.1 & 33.5\\
        LSTM - NER & 26.2 & 42.6 & 32.4\\
        LSTM - KW & 27.4 & 48.0 & 34.9\\
        BERT & 31.0 & \textbf{50.5} & 38.4\\
        \textbf{BERT Predictions \& Features} & \textbf{32.2} & 50.2 & \textbf{39.2}\\
        \hline
        Team syrapropa & 39.9 & 80.8 & 53.4\\
    \end{tabular}
    \caption{SI results on the dev set. Among our models, the best (by F1) is shown in bold. For comparison, we include the dev results for Team syrapropa, who had the best performance on the dev set leaderboard.}
    \label{tab:si_dev_results}
\end{table}

For the test set, we include results for the model that performed the best on the dev set, our BERT Predictions \& Features model. Every model with added features outperforms our original LSTM model, which learns only from fixed BERT embeddings. However, while the part-of-speech and named entity features appear to help, the keyword frequency feature looks like it may actually hurt the model slightly. Removing it lead to roughly a 1\% absolute increase in F1. Our vanilla BERT model outperformed all variants of our LSTM models, as expected, and integrating our BERT model predictions with our features gave us a modest further improvement.

\begin{table}
    \centering\small\setlength\tabcolsep{5pt}
    \begin{tabular} {l | c c c}
        &\textbf{P}&\textbf{R}&\textbf{F}\\
        \hline
        \textbf{\# Gold Spans} & \multicolumn{3}{c}{1791}\\
        \hline

        Our Model & 42.4 & 34.2 & 37.9\\
        Team Hitachi & 56.5 & 47.4 & 51.6\\
    \end{tabular}
    \caption{SI results on the test set. Our model ranks 22 out of 36.}
    \label{tab:si_test_results}
\end{table}

\section{Task 2: Technique Classification}

For TC, we are given propaganda spans annotated according to the set of 14 labels shown in \cref{tab:tc_proportions}. Our approach is similar to our models for SI. We experimented with using an off-the-shelf BERT solution, but failed to find hyperparameters that lead to model convergence. One possible reason for this is the limited size of the training data for the task. Even though the datasets for both tasks are the same, the number of training instances is not, since the two tasks have very different objectives. In SI, every token gets a prediction, whereas in TC, we're only making predictions about previously identified propaganda spans. The training set for TC contains only 6,129 such spans.

\paragraph{Data Augmentation}
Another challenge of this task is the stark imbalance of class labels. In the previous version of this task, several teams addressed the class imbalance by undersampling, and a few took the opposite approach of oversampling. Both tactics lead to better results compared to training directly on the unbalanced dataset. In our approach to the current task, we used the Snorkel library~\footnote{\url{https://github.com/snorkel-team/snorkel}} to add samples to classes with less data (12 of the classes). Snorkel facilitates the creation of ``silver'' data with techniques such as randomly replacing verbs/nouns/adjectives with synonyms and substituting different proper nouns in place of those the original training instances. Using these techniques, we generated about 3,000 new training samples. As shown in \cref{tab:tc_dev_results}, this lead to a relative increase in performance of $4.6\%$.

\subsection{Preprocessing}
All of our models for TC only consider the context of the provided span. In retrospect, incorporating contextual information at the article or sentence level probably should have been a priority, given that some spans can be very short and don't betray much information about themselves on their own.
Preprocessing for TC is much simpler than preprocessing for SI, since we don't have to generate token labels from character indexes. Instead, all we have to do is tokenize the spans and then featurize. We use the same features as described in \cref{sec:si_models}: POS tags, named entities, and KW frequencies.

\subsection{Models}
On the heels of the success we had with featurization in SI, our first attempt at TC is an LSTM over fixed BERT embeddings and all features. Then, we also experiment with training on an expanded training set (the union of the original training set and our previously described augmented data.)

\subsection{Results}
Many patterns in our results are expected, but some stand out. As we might have anticipated, our model did best on the two most frequent classes, which together make up more than 50\% of the training data. Incorporating the augmented data into our training set to address class imbalance helped our model do better on rare classes, as compared to some of the systems with better overall performance.

Curiously, our model is not as good at identifying \textit{Repetition} compared to other classes, despite the fact that this is the third most frequent class in the training data. On the surface, this is surprising, since repetition is usually superficially simple---it's often just multiple uses of the same token. Upon further inspection though, this seems to be an artifact of the annotation schema and the fact that we are not making use of contextual information around the span. In the training data, most spans of repetition are single tokens (likely where the second instance of the token is the one marked as propaganda and the first is unmarked). This creates two challenges: first, since the model only looks at the span as input, it is unaware that an identical token is nearby, and second, the model may learn to falsely associate specific tokens with being propaganda, leading to many false positives when that token is encountered at test time. Incorporating sentence context into the model input could lead to a significant increase in the model's ability to recognize instances of repetition, and potentially other classes that suffer from similar problems.

On flag-waving, our model does remarkably well given that flag-waving makes up only 4\% of the training dataset. Flag-waving is likely an easier class in general, drawing from a smaller vocabulary of nationalistic terms that are easy for the model to pick out, compared to other classes.

And finally, we observe that while the mean span length varies considerably between classes, there doesn't seem to be a clear high-level correlation one way or the other between mean span length and F1.

\begin{table}[h]
    \centering\small\setlength\tabcolsep{5pt}
    \begin{tabular}{l | c c c}
    \textbf{Propaganda Class} & \textbf{Proportion} & \textbf{Mean Span Length} & \textbf{F}\\
    \hline
    Loaded language & 35\% & 24 & 68.4\\
    Name calling, labeling & 17\% & 26 & 60.6\\
    Repetition & 10\% & 18 & 19.4\\
    Doubt & 8\% & 125 & 46.3\\
    Exaggeration, minimisation & 8\% & 44 & 27.2\\
    Appeal to fear/prejudice & 5\% & 99 & 29.8\\
    Flag-waving & 4\% & 62 & 53.8\\
    Causal oversimplification & 3\% & 124 & 14.9\\
    Appeal to authority & 2\% & 139 & 22.6\\
    Slogans & 2\% & 25 & 28.1\\
    Whataboutism, straw man, red herring & $\leq$ 2\% & 97 & 9.7\\
    Black-and-white fallacy & $\leq$ 2\% & 105 & 24.5\\
    Though-terminating clichés & $\leq$ 2\% & 30 & 12.2\\
    Bandwagon, reductio ad hitlerum & $\leq$ 2\% & 96 & 4.5\\
    
    \end{tabular}
    \caption{Test set micro-average F1 scores sorted by class proportion in the unaugmented training set. Mean length of propaganda spans measured in characters.}
    \label{tab:tc_proportions}
\end{table}

\begin{table}[h]
  \centering\small\setlength\tabcolsep{5pt}
  \begin{subtable}[b]{0.4\textwidth}
    \begin{center}
      \begin{tabular}[]{l | c}
        &\textbf{F}\\
        \hline
        \textbf{\# Gold Spans} & \multicolumn{1}{c}{1,064}\\
        \hline
        
        LSTM & 51.7\\
        LSTM + Data Augmentation & 54.1\\
        \textbf{LSTM + Finetuning} & \textbf{54.8}\\
        LSTM + Finetuning + Data Augmentation & 54.5\\
        
        \hline
        Team ApplicaAI & 70.4\\
    \end{tabular}
      \caption{}
      \label{tab:left}
    \end{center}
  \end{subtable}
  \quad
  \begin{subtable}[b]{0.4\textwidth}
    \begin{center}
      \begin{tabular}[]{l | c}
        &\textbf{F}\\
        \hline
        \textbf{\# Gold Spans} & \multicolumn{1}{c}{1,791}\\
        \hline
        Our Model & 49.7\\
        \textbf{Team ApplicaAI} & \textbf{62.1}\\
    \end{tabular}
      \par\bigskip\par\bigskip\par\smallskip\par\smallskip\caption{}
      \label{tab:right}
    \end{center}
    \end{subtable}
  \caption{TC results on the dev set (a) and test set (b). P and R are not shown because this is a multiclass task. F1 is microaveraged. Our model ranks 24 out of 32 on the test set.}
  \label{tab:tc_dev_results}
\end{table}

\section{Conclusion}

We investigated several models and combinations of features to identify propaganda spans in text, and classify the techniques used within the span. For the span identification task, we found that our LSTM-based model combining BERT predictions with our original features gives the highest F1 score of 39.2\% on the dev set. Among the features we used, named entities and POS tags were found to be the most useful for this task. For the technique classification task, when using pre-trained BERT, we found data augmentation helps the LSTM model, however, when we finetuned BERT, we got better results without using the augmented data for training. Our best model for this task got an F1 score of 54.8\% on the dev set.

The top teams in the SemEval 2020 shared task achieved an F1 score of 51.6\% and 62.1\% on the test set for the SI and TC task respectively. These scores represent significant improvements over the baselines. Nonetheless, we believe that there is still room for improvement on this challenging task, and we hope that our results, ablations, discussions, and the release of our models will help the research community in further studies.

\bibliographystyle{coling}

\bibliography{semeval}

\begin{thebibliography}{}

\bibitem[\protect\citename{Aggarwal and Sadana}2019]{aggarwal-sadana-2019-nsit}
Kartik Aggarwal and Anubhav Sadana.
\newblock 2019.
\newblock {NSIT}@{NLP}4{IF}-2019: Propaganda detection from news articles using
  transfer learning.
\newblock In {\em Proceedings of the Second Workshop on Natural Language
  Processing for Internet Freedom: Censorship, Disinformation, and Propaganda}.
  Association for Computational Linguistics.

\bibitem[\protect\citename{Al-Omari \bgroup et al.\egroup
  }2019]{al-omari-etal-2019-justdeep}
Hani Al-Omari, Malak Abdullah, Ola AlTiti, and Samira Shaikh.
\newblock 2019.
\newblock {JUSTD}eep at {NLP}4{IF} 2019 task 1: Propaganda detection using
  ensemble deep learning models.
\newblock In {\em Proceedings of the Second Workshop on Natural Language
  Processing for Internet Freedom: Censorship, Disinformation, and Propaganda}.
  Association for Computational Linguistics.

\bibitem[\protect\citename{Alhindi \bgroup et al.\egroup
  }2019]{alhindi-etal-2019-fine}
Tariq Alhindi, Jonas Pfeiffer, and Smaranda Muresan.
\newblock 2019.
\newblock Fine-tuned neural models for propaganda detection at the sentence and
  fragment levels.
\newblock In {\em Proceedings of the Second Workshop on Natural Language
  Processing for Internet Freedom: Censorship, Disinformation, and Propaganda}.
  Association for Computational Linguistics.

\bibitem[\protect\citename{Baly \bgroup et al.\egroup
  }2018]{baly-etal-2018-predicting}
Ramy Baly, Georgi Karadzhov, Dimitar Alexandrov, James Glass, and Preslav
  Nakov.
\newblock 2018.
\newblock Predicting factuality of reporting and bias of news media sources.
\newblock In {\em Proceedings of the 2018 Conference on Empirical Methods in
  Natural Language Processing}. Association for Computational Linguistics.

\bibitem[\protect\citename{Barr{\'o}n-Cede\~no \bgroup et al.\egroup
  }2019a]{barron2019proppy}
Alberto Barr{\'o}n-Cede\~no, Giovanni Da~San~Martino, Israa Jaradat, and
  Preslav Nakov.
\newblock 2019a.
\newblock Proppy: A system to unmask propaganda in online news.
\newblock In {\em Proceedings of the AAAI Conference on Artificial
  Intelligence}.

\bibitem[\protect\citename{Barr{\'o}n-Cede\~no \bgroup et al.\egroup
  }2019b]{BARRONCEDENO20191849}
Alberto Barr{\'o}n-Cede\~no, Israa Jaradat, Giovanni Da~San~Martino, and
  Preslav Nakov.
\newblock 2019b.
\newblock Proppy: Organizing the news based on their propagandistic content.
\newblock {\em Information Processing \& Management}, 56(5):1849--1864.

\bibitem[\protect\citename{Cer \bgroup et al.\egroup }2018]{cer2018universal}
Daniel Cer, Yinfei Yang, Sheng-yi Kong, Nan Hua, Nicole Limtiaco, Rhomni~St
  John, Noah Constant, Mario Guajardo-Cespedes, Steve Yuan, Chris Tar, et~al.
\newblock 2018.
\newblock Universal sentence encoder.
\newblock {\em arXiv preprint arXiv:1803.11175}.

\bibitem[\protect\citename{Da~San~Martino \bgroup et al.\egroup
  }2019a]{da-san-martino-etal-2019-findings}
Giovanni Da~San~Martino, Alberto Barr{\'o}n-Cede{\~n}o, and Preslav Nakov.
\newblock 2019a.
\newblock Findings of the {NLP}4{IF}-2019 shared task on fine-grained
  propaganda detection.
\newblock In {\em Proceedings of the Second Workshop on Natural Language
  Processing for Internet Freedom: Censorship, Disinformation, and Propaganda}.
  Association for Computational Linguistics.

\bibitem[\protect\citename{Da~San~Martino \bgroup et al.\egroup
  }2019b]{da-san-martino-etal-2019-fine}
Giovanni Da~San~Martino, Seunghak Yu, Alberto Barr{\'o}n-Cede{\~n}o, Rostislav
  Petrov, and Preslav Nakov.
\newblock 2019b.
\newblock Fine-grained analysis of propaganda in news article.
\newblock In {\em Proceedings of the 2019 Conference on Empirical Methods in
  Natural Language Processing and the 9th International Joint Conference on
  Natural Language Processing (EMNLP-IJCNLP)}. Association for Computational
  Linguistics.

\bibitem[\protect\citename{Da~San~Martino \bgroup et al.\egroup
  }2020]{DaSanMartinoSemeval20task11}
Giovanni Da~San~Martino, Alberto Barr\'{o}n-Cede\~no, Henning Wachsmuth,
  Rostislav Petrov, and Preslav Nakov.
\newblock 2020.
\newblock {SemEval}-2020 task 11: Detection of propaganda techniques in news
  articles.
\newblock In {\em Proceedings of the 14th International Workshop on Semantic
  Evaluation}, SemEval 2020.

\bibitem[\protect\citename{Ek and
  Ghanimifard}2019]{ek-ghanimifard-2019-synthetic}
Adam Ek and Mehdi Ghanimifard.
\newblock 2019.
\newblock Synthetic propaganda embeddings to train a linear projection.
\newblock In {\em Proceedings of the Second Workshop on Natural Language
  Processing for Internet Freedom: Censorship, Disinformation, and Propaganda}.
  Association for Computational Linguistics.

\bibitem[\protect\citename{Fadel \bgroup et al.\egroup
  }2019]{fadel-etal-2019-pretrained}
Ali Fadel, Ibraheem Tuffaha, and Mahmoud Al-Ayyoub.
\newblock 2019.
\newblock Pretrained ensemble learning for fine-grained propaganda detection.
\newblock In {\em Proceedings of the Second Workshop on Natural Language
  Processing for Internet Freedom: Censorship, Disinformation, and Propaganda}.
  Association for Computational Linguistics.

\bibitem[\protect\citename{Ferreira~Cruz \bgroup et al.\egroup
  }2019]{ferreira-cruz-etal-2019-sentence}
Andr{\'e} Ferreira~Cruz, Gil Rocha, and Henrique Lopes~Cardoso.
\newblock 2019.
\newblock On sentence representations for propaganda detection: From
  handcrafted features to word embeddings.
\newblock In {\em Proceedings of the Second Workshop on Natural Language
  Processing for Internet Freedom: Censorship, Disinformation, and Propaganda}.
  Association for Computational Linguistics.

\bibitem[\protect\citename{Gupta \bgroup et al.\egroup
  }2019]{gupta-etal-2019-neural}
Pankaj Gupta, Khushbu Saxena, Usama Yaseen, Thomas Runkler, and Hinrich
  Sch{\"u}tze.
\newblock 2019.
\newblock Neural architectures for fine-grained propaganda detection in news.
\newblock In {\em Proceedings of the Second Workshop on Natural Language
  Processing for Internet Freedom: Censorship, Disinformation, and Propaganda}.
  Association for Computational Linguistics.

\bibitem[\protect\citename{Helmstetter and
  Paulheim}2018]{helmstetter2018weakly}
Stefan Helmstetter and Heiko Paulheim.
\newblock 2018.
\newblock Weakly supervised learning for fake news detection on twitter.
\newblock In {\em 2018 IEEE/ACM International Conference on Advances in Social
  Networks Analysis and Mining (ASONAM)}. IEEE.

\bibitem[\protect\citename{Hou and Chen}2019]{hou-chen-2019-caunlp}
Wenjun Hou and Ying Chen.
\newblock 2019.
\newblock {CAU}n{LP} at {NLP}4{IF} 2019 shared task: Context-dependent {BERT}
  for sentence-level propaganda detection.
\newblock In {\em Proceedings of the Second Workshop on Natural Language
  Processing for Internet Freedom: Censorship, Disinformation, and Propaganda}.
  Association for Computational Linguistics.

\bibitem[\protect\citename{Hua}2019]{hua-2019-understanding}
Yiqing Hua.
\newblock 2019.
\newblock Understanding {BERT} performance in propaganda analysis.
\newblock In {\em Proceedings of the Second Workshop on Natural Language
  Processing for Internet Freedom: Censorship, Disinformation, and Propaganda}.
  Association for Computational Linguistics.

\bibitem[\protect\citename{Kuru \bgroup et al.\egroup
  }2016]{kuru-etal-2016-charner}
Onur Kuru, Ozan~Arkan Can, and Deniz Yuret.
\newblock 2016.
\newblock {C}har{NER}: Character-level named entity recognition.
\newblock In {\em Proceedings of {COLING} 2016, the 26th International
  Conference on Computational Linguistics: Technical Papers}, pages 911--921,
  Osaka, Japan, December. The COLING 2016 Organizing Committee.

\bibitem[\protect\citename{Li \bgroup et al.\egroup
  }2019]{li-etal-2019-detection}
Jinfen Li, Zhihao Ye, and Lu~Xiao.
\newblock 2019.
\newblock Detection of propaganda using logistic regression.
\newblock In {\em Proceedings of the Second Workshop on Natural Language
  Processing for Internet Freedom: Censorship, Disinformation, and Propaganda}.
  Association for Computational Linguistics.

\bibitem[\protect\citename{Mapes \bgroup et al.\egroup
  }2019]{mapes-etal-2019-divisive}
Norman Mapes, Anna White, Radhika Medury, and Sumeet Dua.
\newblock 2019.
\newblock Divisive language and propaganda detection using multi-head attention
  transformers with deep learning {BERT}-based language models for binary
  classification.
\newblock In {\em Proceedings of the Second Workshop on Natural Language
  Processing for Internet Freedom: Censorship, Disinformation, and Propaganda}.
  Association for Computational Linguistics.

\bibitem[\protect\citename{Potthast \bgroup et al.\egroup
  }2018]{potthast-etal-2018-stylometric}
Martin Potthast, Johannes Kiesel, Kevin Reinartz, Janek Bevendorff, and Benno
  Stein.
\newblock 2018.
\newblock A stylometric inquiry into hyperpartisan and fake news.
\newblock In {\em Proceedings of the 56th Annual Meeting of the Association for
  Computational Linguistics (Volume 1: Long Papers)}. Association for
  Computational Linguistics.

\bibitem[\protect\citename{Rashkin \bgroup et al.\egroup
  }2017]{rashkin-etal-2017-truth}
Hannah Rashkin, Eunsol Choi, Jin~Yea Jang, Svitlana Volkova, and Yejin Choi.
\newblock 2017.
\newblock Truth of varying shades: Analyzing language in fake news and
  political fact-checking.
\newblock In {\em Proceedings of the 2017 Conference on Empirical Methods in
  Natural Language Processing}. Association for Computational Linguistics.

\bibitem[\protect\citename{Tayyar~Madabushi \bgroup et al.\egroup
  }2019]{tayyar-madabushi-etal-2019-cost}
Harish Tayyar~Madabushi, Elena Kochkina, and Michael Castelle.
\newblock 2019.
\newblock Cost-sensitive {BERT} for generalisable sentence classification on
  imbalanced data.
\newblock In {\em Proceedings of the Second Workshop on Natural Language
  Processing for Internet Freedom: Censorship, Disinformation, and Propaganda}.
  Association for Computational Linguistics.

\bibitem[\protect\citename{Vlad \bgroup et al.\egroup
  }2019]{vlad-etal-2019-sentence}
George-Alexandru Vlad, Mircea-Adrian Tanase, Cristian Onose, and
  Dumitru-Clementin Cercel.
\newblock 2019.
\newblock Sentence-level propaganda detection in news articles with transfer
  learning and {BERT}-{B}i{LSTM}-capsule model.
\newblock In {\em Proceedings of the Second Workshop on Natural Language
  Processing for Internet Freedom: Censorship, Disinformation, and Propaganda}.
  Association for Computational Linguistics.

\bibitem[\protect\citename{Yoosuf and Yang}2019]{yoosuf-yang-2019-fine}
Shehel Yoosuf and Yin Yang.
\newblock 2019.
\newblock Fine-grained propaganda detection with fine-tuned {BERT}.
\newblock In {\em Proceedings of the Second Workshop on Natural Language
  Processing for Internet Freedom: Censorship, Disinformation, and Propaganda}.
  Association for Computational Linguistics.

\end{thebibliography}

\end{document}